\documentclass[10pt,twocolumn,letterpaper]{article}
\usepackage{cvpr}
\usepackage{amsmath}
\usepackage{amssymb}
\usepackage{booktabs}
\usepackage{tabularx}
\usepackage[normalem]{ulem}
\usepackage[dvipsnames]{xcolor}
\usepackage[accsupp]{axessibility}
\usepackage{enumitem,kantlipsum}
\usepackage[pagebackref,breaklinks,colorlinks]{hyperref}
\usepackage[capitalize]{cleveref}
\usepackage[pdftex]{graphicx}
\crefname{section}{Sec.}{Secs.}
\Crefname{section}{Section}{Sections}
\Crefname{table}{Table}{Tables}
\crefname{table}{Tab.}{Tabs.}
\Crefname{equation}{Equation}{Equations}
\crefname{equation}{eq.}{eqs.}

\newcommand\CA[1]{\textcolor{teal}{#1}}
\newcommand\CAcr[1]{\textcolor{teal}{#1}}
\newcommand\CAnew[1]{#1}

\renewcommand\CA[1]{\textcolor{black}{#1}}
\renewcommand\CAcr[1]{\textcolor{black}{#1}}

\def\cvprPaperID{9972}
\def\confName{CVPR}
\def\confYear{2023}

\makeatletter
\renewcommand{\paragraph}{%
  \@startsection{paragraph}{4}%
  {\z@}{0.25em}{-1em}%
  {\normalfont\normalsize\bfseries}%
}
\makeatother

\begin{document}

\title{Novel-View Acoustic Synthesis}

\author{
Changan Chen$^{1,3}$ \hspace{3mm} Alexander Richard$^{2}$ \hspace{3mm} Roman Shapovalov$^{3}$ \hspace{3mm} Vamsi Krishna Ithapu$^{2}$ \hspace{3mm}\\
Natalia Neverova$^{3}$ \hspace{3mm} Kristen Grauman$^{1,3}$ \hspace{3mm} Andrea Vedaldi$^{3}$  \smallskip \\
$^1$University of Texas at Austin \hspace{3mm}$^2$Reality Labs Research at Meta \hspace{3mm} $^3$FAIR, Meta AI
}

\maketitle

\begin{abstract}
We introduce the novel-view acoustic synthesis (NVAS) task: given the sight and sound observed at a source viewpoint, can we synthesize the \emph{sound} of that scene from an unseen target viewpoint? We propose a neural rendering approach: Visually-Guided Acoustic Synthesis (ViGAS) network that learns to synthesize the sound of an arbitrary point in space by analyzing the input audio-visual cues. 
To benchmark this task, we collect two first-of-their-kind large-scale multi-view audio-visual datasets, one synthetic and one real. 
We show that our model successfully reasons about the spatial cues and synthesizes faithful audio on both datasets. To our knowledge, this work represents the very first formulation, dataset, and approach to solve the novel-view acoustic synthesis task, which has exciting potential applications ranging from AR/VR to art and design.
Unlocked by this work, we believe that the future of novel-view synthesis is in multi-modal learning from videos.
\end{abstract}
\vspace{-0.05in}
\section{Introduction}\label{sec:intro}
\vspace{-0.05in}

Replaying a video recording from a new viewpoint\footnote{We use ``viewpoint'' to mean a camera or microphone pose.} has many applications in cinematography, video enhancement, and virtual reality.
For example, it can be used to edit a video, simulate a virtual camera,
or, given a video of a personal memory, even enable users to experience a treasured moment again---not just on a 2D screen, but in 3D in a virtual or augmented reality, thus `reliving' the moment.

While the applications are exciting, there are still many unsolved technical challenges.
Recent advances in 3D reconstruction and novel-view synthesis (NVS) address the problem of synthesizing new \emph{images} of a given scene~\cite{mildenhall2020nerf,Martin-Brualla_2021_CVPR,Pumarola_2021_CVPR}.
However, thus far, the view synthesis problem is concerned with creating visuals alone; the output is silent or at best naively adopts the sounds of the original video (from the ``wrong" viewpoint).
Without sound, the emotional and cognitive significance of the replay is severely diminished.

\begin{figure}[t]
\centering
\includegraphics[width=\linewidth]{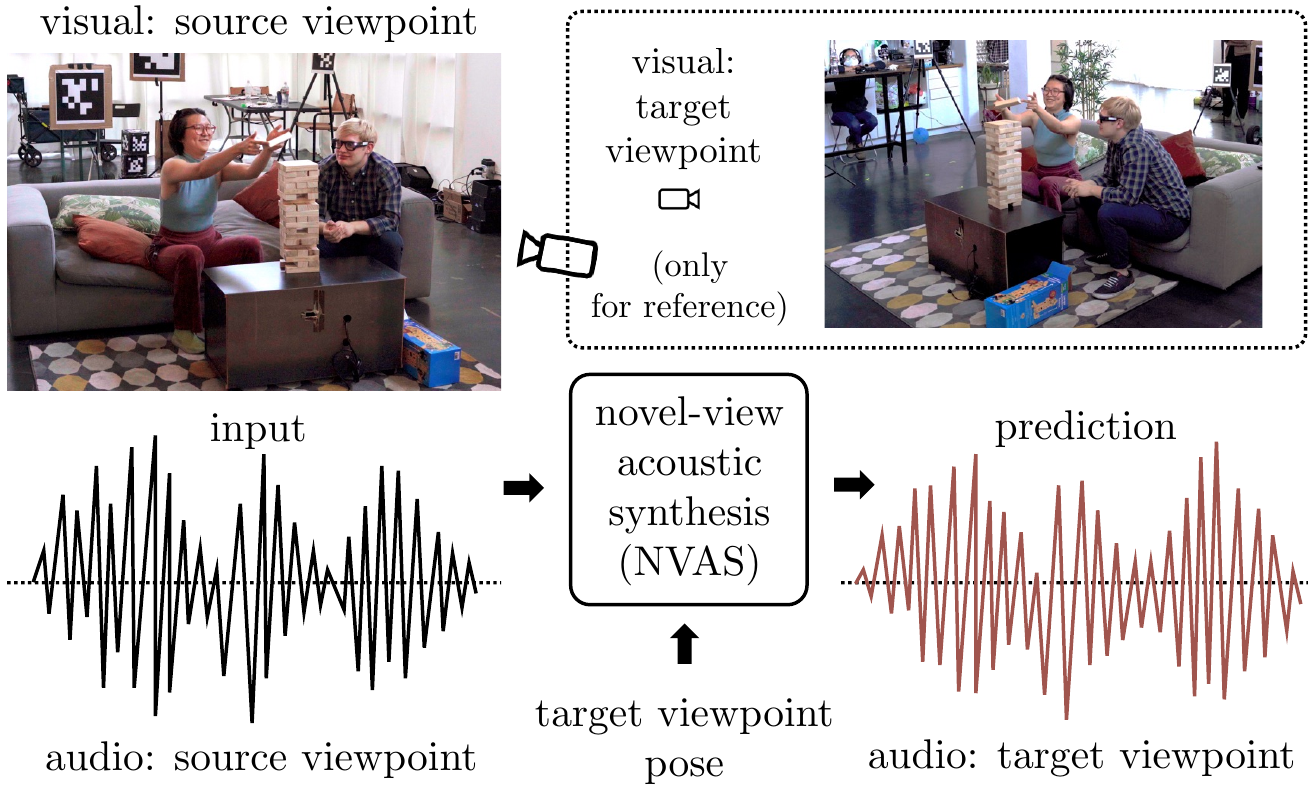}
\vspace{-0.2in}
\caption{\textbf{Novel-view acoustic synthesis task.} Given audio-visual observations from one viewpoint and the relative target viewpoint pose, render the sound received at the target viewpoint.  Note that the target is expressed as the desired pose of the microphones; the image at that pose (right) is neither observed nor synthesized.
}\label{fig:concept}
\vspace{-0.1in}
\end{figure}

In this work, we address this gap and introduce the new task of \emph{novel-view acoustic synthesis} (NVAS).
The goal of this task is to synthesize the sound in a scene from a new acoustic viewpoint, given only the visual and acoustic input from another source viewpoint in the same scene (Fig.~\ref{fig:concept}). 

NVAS is very different from the existing NVS task, where the goal is to reconstruct images instead of sounds, and these differences present new challenges.
First, the 3D geometry of most real-life scenes changes in a limited manner during the recording.
On the contrary, sound changes substantially over time, so the reconstruction target is highly dynamic.
Secondly, visual and audio sensors are very different.
A camera matrix captures the light in a highly-directional manner, and a single image comprises a large 2D array of pixels.
In contrast, sounds are recorded with one or two microphones which are at best weakly-directional, providing only a coarse sampling of the sound field.
Thirdly, the frequency of light waves is much higher than that of sound waves; the length of audio waves is thus larger to the point of being comparable to the size of geometric features of the scene, meaning that effects such as diffraction are often dominant, and spatial resolution is low.
As a result, techniques that require spatial precision, such as triangulation and segmentation, are not applicable to audio.
Lastly, sounds mix together, making it difficult to segment them, and they are affected by environmental effects such as reverberation that are distributed and largely unobservable.

While the NVS and NVAS tasks are indeed very different, we hypothesize that NVAS is an inherently multimodal task.  In fact, vision can play an important role in achieving accurate sound synthesis.
First, establishing correspondences between sounds and their sources as they appear in images can provide essential cues for resynthesizing the sounds realistically.
For instance, human speech is highly directional and sounds very differently if one faces the speaker or their back, which can only be inferred from visual cues.
In addition, the environment acoustics also affect the sound one hears as a function of the scene geometry, materials, and emitter/receiver locations.
The same source sounds very differently if it is located in the center of a room, at the corner, or in a corridor, for example.
In short, vision provides cues about space and geometry that affect sound, and are difficult to estimate from the sound alone.

In order to validate our hypothesis, we propose a novel \emph{visually-guided acoustic synthesis network} that analyzes audio and visual features and synthesizes the audio at a target location.
More specifically, the network first takes as input the image observed at the source viewpoint in order to infer global acoustic and geometric properties of the environment along with the  
bounding box of the active speaker. 
\CA{The network then reasons how the speaker and scene geometry change in 3D based on the relative target pose with a fusion network.}
\CA{We inject the fused features into audio with a gated multi-modal fusion network and model the acoustic changes between viewpoints with a time-domain model.}

In order to conduct our experiments on the new NVAS task, we require suitable training and benchmarking data, of which currently there is none available.
To address that, we contribute two new datasets: one real (Replay-NVAS) and one synthetic (SoundSpaces-NVAS).
The key feature of these datasets is to record the sight and sound of different scenes from multiple cameras/viewpoints. 
Replay-NVAS contains video recordings of groups of people performing social activities (e.g., chatting, watching TV, doing yoga, playing instruments) from 8 surrounding viewpoints simultaneously.  
It contains \CAcr{37} hours of highly realistic everyday conversation and social interactions in one home-like environment.  
To our knowledge, Replay-NVAS represents the first large-scale real-world dataset enabling NVAS. 
\CA{This dataset would also greatly benefit many other existing tasks including NVS, active speaker localization, etc.}
For SoundSpaces-NVAS, we render 1.3K hours of audio-visual data based on the SoundSpaces~\cite{chen22soundspaces2} platform.
Using this simulator, one can easily change the scene geometry and the positions of speakers, cameras, and microphones.
This data serves as a 
powerful test bed with clean ground truth for a large collection of home environments, offering a good complement to Replay-NVAS.
For both datasets, we capture binaural audio, which is what humans perceive with two ears.
Together the datasets contain \CAcr{1,337} hours of audio-visual capture, with 1,032 speakers across 121 3D scenes. 
\CAcr{Datasets are publicly available for future research.}~\footnote{\url{https://replay-dataset.github.io}}

We show that our model outperforms traditional signal processing approaches as well as learning-based baselines, often by a substantial margin, in a quantitative evaluation and a human study.
We show qualitative examples where the model predicts acoustic changes according to the viewpoint changes, \CA{e.g., left channel becomes louder when the viewpoint changes from left to right}.
In a nutshell, we present the first work that deals with novel-view acoustic synthesis, and contribute two large-scale datasets along with a novel neural rendering approach for solving the task.

\vspace{-0.05in}
\section{Related Work}%
\label{sec:related_work}
\vspace{-0.05in}
\paragraph{Novel-view synthesis (NVS).}

Kickstarted by advances in neural rendering~\cite{sitzmann19scene,mildenhall2020nerf}, many recent works consider variants of the NVS problem.
Most approaches assume dozens of calibrated images for reconstructing a single static scene.
Closer to monocular video NVS, authors have considered reducing the number of input views~\cite{niemeyer22regnerf:,jain21putting,reizenstein21common,yu21pixelnerf:,kulhanek22viewformer:} and modelling dynamic scenes~\cite{li20neural,pumarola20d-nerf:,park20deformable,liu21neural,su21a-nerf:,tretschk20non-rigid}.
However, none of these works tackle audio.

\paragraph{Acoustic matching and spatialization.}

NVAS requires accounting for (1) the environmental acoustics and (2) the geometric configuration of the target microphone(s) (e.g., monaural vs binaural).
Modelling environmental acoustics has been addressed extensively by the audio community~\cite{room_acoustics,room_acoustics_review}. 
Room impulse response (RIR) functions characterize the environment acoustics as a transfer function between the emitter and receiver, accounting for the scene geometry, materials, and emitter/receiver locations.
Estimating the direct-to-reverberant ratio
and the reverberation time, 
is sufficient to synthesize simple RIRs that match audio in a plausible manner~\cite{eaton20161ACE, gamper2018blind, klein2019real, mack2020single, murgai2017blind, xiong2018joint}.  These methods do not synthesize for a target viewpoint, rather they resynthesize to match an audio sample.
In \cite{richard2021binaural,richard2022deepimpulse} sound from a moving emitter is spatialized towards a receiver conditioned on the tracked 3D location of the emitter. 

Recently, the vision community 
explores using visual information to estimate environmental acoustics~\cite{singh_image2reverb_2021,chen2022vam,chen_dereverb23}.
However, these works only synthesize acoustics for a given viewpoint rather than a novel viewpoint.
In addition, they have only addressed monaural audio, which is more forgiving than binaural because humans do not perceive subtle absolute acoustic properties, but can detect easily inconsistencies in the sounds perceived by the two ears.
Recent work 
spatializes monaural sounds by upmixing them to multiple channels conditioned on the video, where the sound emitters are static~\cite{25d-visual-sound,morgado-2018}.
Because the environment, emitter and receiver are static, so are the acoustics.
Other work predicts impulse responses in simulation either for a single environment~\cite{luo2022learning}, or by using few-shot egocentric observations~\cite{majumder2022fewshot}, or by using the 3D scene mesh~\cite{ratnarajah2022mesh2ir}.
While simulated results are satisfying, 
those models' impact on real-world data is unknown,
especially for scenarios where human speakers move and interact with each other.
Unlike any of the above, we introduce and tackle the NVAS problem, accounting for both acoustics and spatialization, and we propose a model that addresses the problem 
effectively on both synthetic and real-world data.

\paragraph{Audio-visual learning.}

Recent advances in multi-modal video understanding 
enable new forms of self-supervised cross-modal feature learning from video~\cite{lorenzo-nips2020,korbar-nips2018,morgado-spatial-nips2020}, sound source localization~\cite{hu-localize-nips2020,hao2022asl,tao2021someone}, and audio-visual speech enhancement and source separation~\cite{ephrat2018looking,owens2018audio,Afouras18,Michelsanti19,zhou19}. 
All of these existing tasks and datasets only deal with a single viewpoint. We introduce the first audio-visual learning task and dataset that deals with multi-view audio-visual data.
%\vspace{-0.05in}
\section{The Novel-view Acoustic Synthesis Task}
\label{sec:task}
%\vspace{-0.05in}

We introduce a new task, \textit{novel-view acoustic synthesis} (NVAS).
Assuming there are $N$ sound emitters in the scene (emitter $i$ emits \CAcr{sound} $C^i$ from location $L^i$), given the audio $A_S$ and video $V_S$ observed at the source viewpoint $S$, the goal is to synthesize the audio $A_T$ at the target viewpoint~$T$, as it would sound from the target location, specified by the relative pose $P_T$ of the target microphone (translation and orientation) with respect to the source view (Fig.~\ref{fig:concept}).
Furthermore, we assume that the active sound emitters in the environment are visible in the source camera, but we make no assumptions about the camera at the target location.

The sound at any point $R$ \CAcr{is a function of the space:}
\vspace{-0.1in}
\begin{equation}
    A_R = \mathcal{F}(L^{1,...,N}, C^{1,...,N}, R \mid E),
\vspace{-0.05in}
\end{equation}\label{eq:propagation}%
where $R$ is the receiver location ($S$ or $T$) and $E$ is the environment. The emitted sounds $C^i$ are not restricted to speech but can be ambient noise, sounding objects, etc.
Our goal here is to learn a transfer function $\mathcal{T}(\cdot)$ defined as $A_T = \mathcal{T}(A_S, V_S, P_T)$, where $S,T,L^{1,...,N},C^{1,...,N},E$ are not directly \CA{given} and need to be inferred from $V_S$ and $P_T$, \CA{which makes the task inherently multi-modal.}

\CA{This task is challenging because the goal is to model the sound field of a dynamic scene and capture acoustic changes between viewpoints given one pair of audio-visual measurements. 
While traditional signal processing methods can be applied, we show in Sec.~\ref{sec:experiment} that they perform poorly. In this work, we present a \CAcr{learning-based} rendering approach.}
%\vspace{-0.05in}
\section{Datasets}\label{sec:datasets}
%\vspace{-0.05in}

\begin{figure}
\centering
\includegraphics[width=0.90\linewidth]{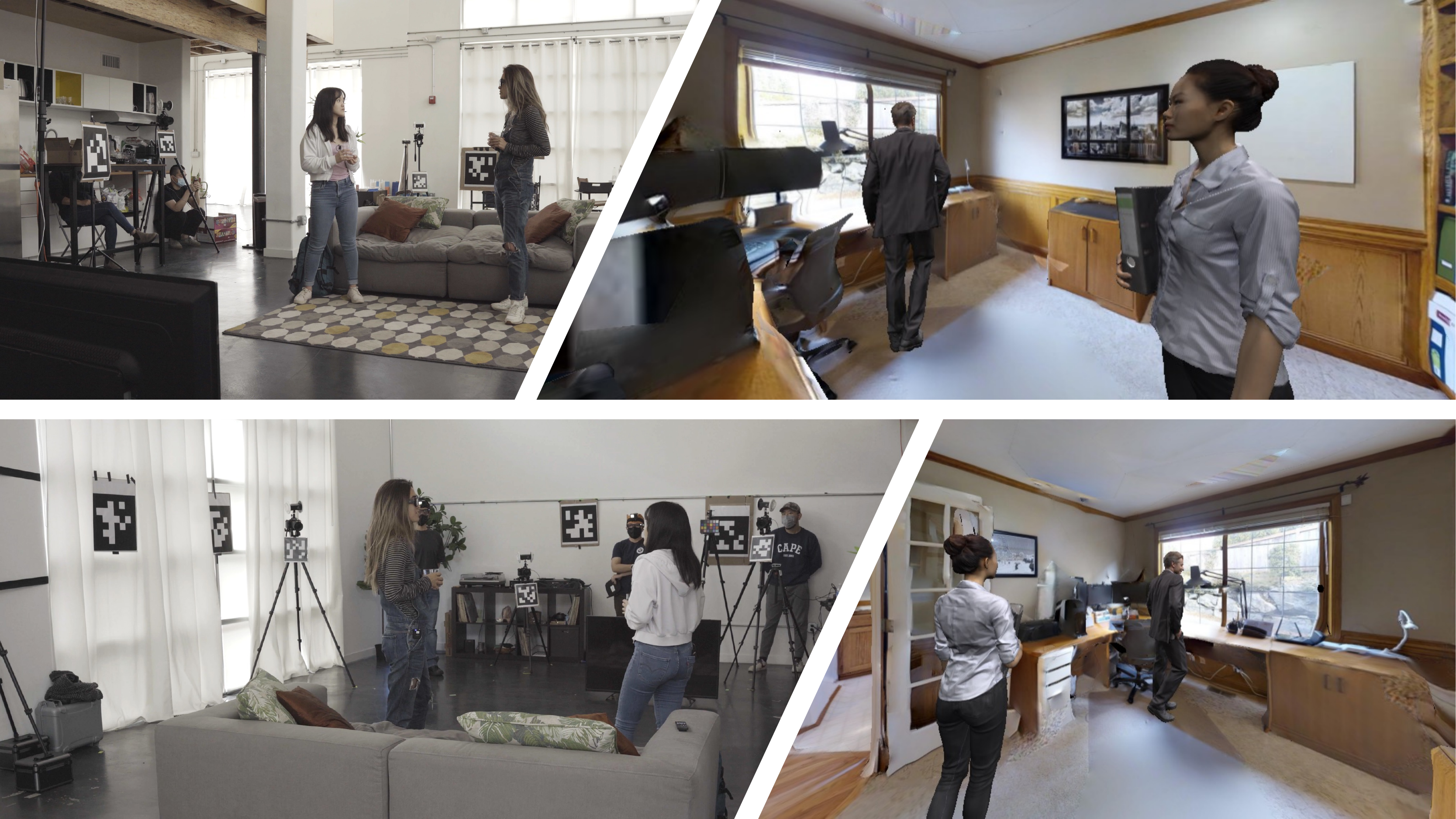}
\vspace{-0.1in}
\caption{Example source and target views for the two introduced datasets: Replay-NVAS (left) and SoundSpaces-NVAS (right).}%
\vspace{-0.2in}
\label{fig:dataset}
\end{figure}

We introduce two datasets for the NVAS task:
live recordings (\cref{sec:replay_dataset}), and simulated audio in scanned real-world environments (\cref{sec:soundspace_nas_dataset}) (see \cref{fig:dataset}).
The former is real \CA{and covers various social scenarios}, but offers limited diversity of sound sources, \CAnew{viewpoints} and environments, and is noisy.
The latter has a realism gap, but allows perfect control over \CAnew{these aforementioned elements}.

Both datasets focus on human speech given its relevance in applications.
However, our model design is not specific to speech.
\CA{For both datasets, we capture binaural audio, which best aligns with human perception.}
Note that for both datasets, we collect \CA{multiple} multi-modal views for training and evaluation; during inference the target viewpoint(s) (and in some cases target environment) are withheld.
% We will release both datasets to assist future research.

%\vspace{-0.05in}
\subsection{The Replay-NVAS Dataset}\label{sec:replay_dataset}
%\vspace{-0.05in}

Replay-NVAS contains multi-view captures of acted scenes in apartments.
We capture \CAcr{46} different scenarios (e.g., having a conversation, having dinner, or doing yoga) from 8 different viewpoints.
In total, we collect \CAcr{37} hours of video data, involving 32 participants across all scenarios.

In each scenario, we invite 2--4 participants to act on a given topic.
Each participant wears a near-range microphone, providing a clean recording of their own speech.
The scene is captured by 8 DLSR cameras, each augmented with a 3Dio binaural microphone.
In this way, the data captures video and audio simultaneously from multiple cameras, resulting in 56 possible source/target viewpoint combinations for each scene.
The videos are recorded at 30 FPS and the audio is recorded with a 48k sampling rate.
We use a clapper at the beginning of the recording for temporal synchronization.
Each scenario lasts 3--8 min.
We use off-the-shelf software for multi-view camera calibration (see Supp.).

To construct the dataset, we extract one-second long clips from each video with overlapping windows. We automatically remove silent and noisy clips based on the energy of near-range microphones, which results in 77K/12K/2K clips in total for train/val/test (details in Supp.)
During training, \CA{for one sample,} we randomly select two out of eight viewpoints, one as the source and one as the target. 

This dataset is very challenging.
It covers a wide range of social activities.
It is harrowed by ambient sound, room reverberation, overlapping speech and non-verbal sounds such as clapping and instruments.
Participants can move freely in the environment.
We believe that this data will be useful to the community beyond the NVAS task as it can be used for benchmarking many other problems, including active speaker localization, source separation, and NVS\@.

\vspace{-0.05in}
\subsection{The SoundSpaces-NVAS Dataset}\label{sec:soundspace_nas_dataset}
\vspace{-0.05in}

In this dataset, we synthesize multi-view audio-visual data of two people having conversations in 3D scenes. In total, we construct 1.3K hours of audio-visual data for a total of 1,000 speakers, 120 3D scenes and 200K viewpoints.

Our goal is to construct audio-visual data with strong spatial and acoustic correspondences across multiple viewpoints, meaning that the visual information should indicate what the audio should sound like, e.g., observing speaker on the left should indicate the left ear is louder and observing speaker at a distance should indicate there is higher reverberation.
We use the SoundSpaces 2.0 platform~\cite{chen22soundspaces2}, which allows highly realistic audio and visual rendering for arbitrary camera and microphone locations in 3D scans of real-world environments~\cite{Matterport3D,straub2019replica,xiazamirhe2018gibsonenv}. 
It accounts for all major real-world acoustics phenomena: direct sounds, early specular/diffuse reflections, reverberation, binaural spatialization, and effects from materials and air absorption.

We use the Gibson dataset~\cite{xiazamirhe2018gibsonenv} for scene meshes and LibriSpeech~\cite{librispeech} for speech samples.
As we are simulating two people having conversations, for a given environment, we randomly sample two speaker locations within 3 m and insert two copyright-free mannequins (one male and one female) at these two locations.\footnote{\url{https://renderpeople.com/free-3d-people}} 
We then randomly sample four nearby viewpoints facing the center of the two speakers at a height of 1.5 m (\cref{fig:dataset}, right). 
For each speaker, we select a speech sample from LibriSpeech with matching gender.
We render images at all locations as well as binaural impulse response for all pairs of points between speakers and viewpoints. 
The received sound is obtained by convolving the binaural impulse response with the speech sample.

During training, \CA{for one sample,} we randomly sample two out of four \CA{rendered viewpoints}, one as the source and one as the target.
We also randomly choose one speaker to be active, \CAnew{simulating what we observe on the real data (i.e., usually only one person speaks at a time)}. 

%\vspace{-0.05in}
\section{Visually-Guided Acoustic Synthesis}%
\label{sec:approach}
%\vspace{-0.05in}

We introduce a new method, \textbf{Vi}sually-\textbf{G}uided \textbf{A}coustic \textbf{S}ynthesis (ViGAS), to address 
the NVAS problem, taking as input sound and an image and outputting the sound from a different target microphone pose.

\begin{figure*}[ht]
\centering
\includegraphics[width=0.95\linewidth]{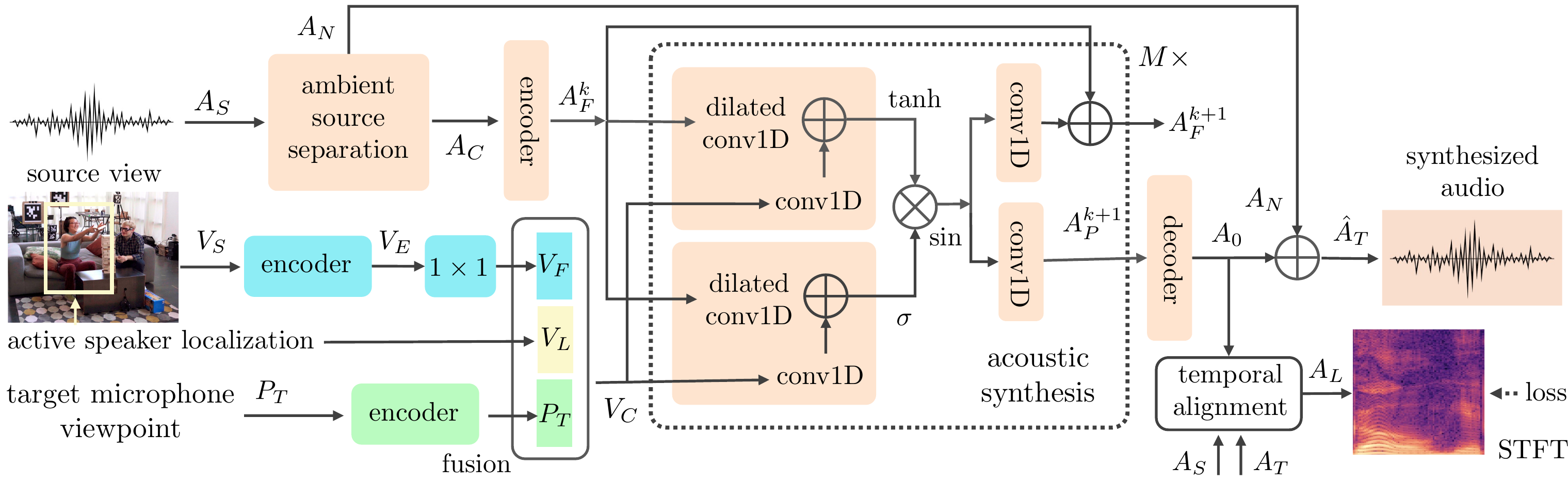}
\vspace{-0.05in}
\caption{\textbf{Visually Guided Acoustic Synthesis (ViGAS)}.
Given the input audio $A_S$, we first separate out the ambient sound to focus on the sound of interest.
We take the source audio and source visual to localize the active speaker on the 2D image.
We also extract the visual acoustic features of the environment by running an encoder on the source visual.
We concatenate the active speaker feature, source visual features, and the target pose, and fuse these features with a MLP.
We feed both the audio stream $A_C$ and fused visual feature $V_C$ into the acoustic synthesis network, which has $M$ stacked audio-visual fusion blocks.
In each block, the audio sequence is processed by dilated conv1d layers and the visual features are processed by conv1d layers.
Lastly, the previously separated ambient sound is added back to the waveform.
During training, our temporal alignment module shifts the prediction by the amount of delay estimated between the source and the target audio to align the prediction well with the target. 
}
\vspace{-0.2in}
\label{fig:approach}
\end{figure*}

ViGAS consists of five components: ambient sound separation, active speaker localization, visual acoustic network, acoustic synthesis, and temporal alignment. 
The high-level idea is to separate the observed sound into primary and ambient, extract useful visual information (active speaker and acoustic features), and use this information to guide acoustic synthesis for the primary sound. 
Temporal alignment is performed during training for better optimization. ViGAS is discussed in detail next and summarised in \cref{fig:approach}.

%\vspace{-0.05in}
\subsection{Ambient Sound Separation}
%\vspace{-0.05in}

ViGAS starts by decomposing the input sound into primary and ambient (traffic, electric noise from a fridge or the A/C, etc.).
Ambient sound is important for realism, but it also interferes with learning the model because
it can carry significant energy, making the model focus on it rather than on the primary sounds, and
its spatial distribution is very different from the primary sounds.

By explicitly separating primary and ambient sounds, ViGAS\@:
(1)
accounts for the fact that the transfer functions of primary and ambient sounds are very different and thus difficult to model together;
(2)
avoids wasting representational power on modelling ambient sounds that might be difficult to reconstruct accurately and depend less on the viewpoint; and
(3)
prevents ambient sounds, which are noise-like and high-energy, from dominating learning and reconstruction.
In practice, as we show in \cref{sec:experiment}, without the ambient sound separation, the model performs poorly.

The goal of ambient sound separation is thus to construct a function $(A_C, A_N)=\mathcal{P}(A_S)$ that separates the input sound $A_S$ into primary sound $A_C$ and ambient sound $A_N$.
Existing approaches to this problem are based on signal processing~\cite{subspace1995,acoustic1979} or learning~\cite{defossez2020real,fu2019metricGAN}.
We find that pretrained speech enhancement models such as Denoiser~\cite{defossez2020real} tend to aggressively remove the noise including the primary sound, which hinders re-synthesis.
We thus opt for band-pass filtering, passing frequencies within a certain range and rejecting/attenuating frequencies outside of it, which we found to work well.
We cut frequencies below 80\,Hz for SoundSpaces-NVAS and 150\,Hz for Replay-NVAS.

%\vspace{-0.05in}
\subsection{Active Speaker Localization}
%\vspace{-0.05in}

Knowing where the emitters of different primary sounds are located in the environment can help to solve the NVAS task.
In this paper, we focus on localizing the active speaker, although there can be other important primary sound events like instruments playing, 
speakers interacting with objects, etc.
The goal of active speaker localization is to predict the bounding box of the active speaker in each frame of the video (examples in Fig.~\ref{fig:qual}). The bounding box is in the format of $(y_{\min}, y_{\max}, x_{\min}, x_{\max})$ and $x,y$ are normalized to $[0, 1]$ by the image width and height, respectively.

On SoundSpaces-NVAS, this task is relatively easy because of the strong correspondence between the appearance of the speaker and the gender of the speech sample, which enables to easily train a classifier for active speakers.
However, this is much harder on Replay-NVAS because cameras record speakers from a distance and from diverse angles, meaning that lip motion, the main cue used by speaker localization methods~\cite{hao2022asl,tao2021someone,ava_active_speaker}, is often not visible.
Hence, the model has to rely on other cues to identify the speaker (such as body motion, gender or identity).
Furthermore, sometimes \CA{people speak} or laugh over each other.

Since our focus is not speaker localization, for the Replay-NVAS we assume that this problem is solved by an external module that does audio-visual active speaker localization.
To approximate the output of such a module automatically, we rely on the near-range audio recordings.
Specifically, we first run an off-the-shelf detection and tracker~\cite{mmtrack2020} on the video at 5 FPS and obtain, with some manual refinement, bounding boxes $B^i_t$ for $i=1,\dots,N$ at each frame $t$.
We manually assign the near-range microphone audio $A_N^i$ to each tracked person.
We select the active speaker $D$ based on the maximum energy of each near-range microphone, i.e.,
$
D = \operatornamewithlimits{argmax}_i
\big\{\sum A_N^i[t: t + \Delta t]^2\big\},
$
where $\Delta t$ is the time interval we use to calculate the audio energy. 
\CA{We output bounding box $B^D$ as the localization feature $V_L$.}

%\vspace{-0.05in}
\subsection{Visual Acoustic Network and Fusion}
%\vspace{-0.05in}

The active speaker bounding box $B^D$ only disambiguates the active speaker from all visible humans on 2D, which is not enough to indicate where the speaker is in 3D. To infer that, the visual information is also needed.  
Since there is usually not much movement in one second (the length of the input video clip), the video clip does not provide much extra information compared to a single frame. 
Thus, we choose the middle frame to represent the clip and extract the visual acoustic features $V_E$ from the input RGB image with a pretrained ResNet18~\cite{resnet18} before the average pooling layer to preserve spatial information. To reduce the feature size, we feed $V_E$ into a 1D convolution with kernel size 1 and output channel size 8. We then flatten the visual features to obtain feature $V_F$.

The target pose is specified as the translation along $x,y,z$ axes plus difference between orientations of 
the source ``view'' and the target ``view" expressed via rotation angles:
$+y$ (roll), $+x$ (pitch) and $+z$ (yaw). We encode each angle $\alpha$ as its sinusoidal value: $(\sin(\alpha), \cos(\alpha))$.

Similarly, the target pose is not enough by itself to indicate where the target viewpoint $T$ is in the 3D space; to infer that, the \CAnew{source view $V_S$} is again needed.
\CAnew{For example, in top row of Fig~\ref{fig:qual}, for target viewpoint 3,  ``two meters to the right and one meter forward'' is not enough to indicate the target location is in the corridor, while the model can reason that based on the source view.}

We use a fusion network to predict a latent representation of the scene variables $S,T,L^D,E$ (\cf Sec.~\ref{sec:task}) by first concatenating $[V_L, P_T, V_F]$ and then feeding it through a multilayer perceptron (MLP). 
\CAnew{See \cref{fig:approach} for the network.}

%\vspace{-0.05in}
\subsection{Acoustic Synthesis}
%\vspace{-0.05in}
With the separated primary sound $A_C$ and the visual acoustic feature $V_C$ as input, the goal of the acoustic synthesis module is to transform $A_C$ guided by~$V_C$. 
We design the acoustic synthesis network to learn a non-linear transfer function (implicitly) that captures these major acoustic phenomena, including the attenuation of sound in space, the directivity of sound sources \CA{(human speech is directional)}, the reverberation level, the head-related transfer function, as well as the frequency-dependent acoustic phenomena. 
Training end-to-end makes it possible to capture these subtle and complicated changes in the audio.

Inspired by recent advances in time-domain signal modeling~\cite{oord2016wavenet,richard2021binaural}, we design the network as $M$ stacked synthesis blocks, where each block consists of multiple conv1D layers. 
We first encode the input audio $A_C$ into a latent space, which is then fed into the synthesis block. The key of the synthesis block is a gated multimodal fusion network that injects the visual information into the audio as follows:
\vspace{-0.2in}
\begin{equation}
    z \!=\! \tanh(p_A^k(A_F^k) \!+\! p_V^k(V_C)) \!\odot\! \sigma(q_A^k(A_F^k) \!+\! q_V^k(V_C)),
\end{equation}
where $\odot$ indicates element-wise multiplication, $\sigma$ is a logistic sigmoid function, $k=1,\dots,M$ is the layer index and $p,q$ are both learnable 1D convolutions.

After passing $z$ through a sinusoidal activation function, the network uses two separate conv1D layers to process the feature, one producing the residual connection $A_F^{k+1}$ and one producing the skip connection $A_P^{k+1}$. All skip connections $A_P^{k+1}$ are mean pooled and fed into a decoder to produce the output $A_O$. We add back the separated ambient sound $A_N$ as the target audio estimate: $\hat{A}_T = A_O + A_N$.
\CAnew{See Supp. for more details on the architecture.}

%\vspace{-0.05in}
\subsection{Temporal Alignment}
%\vspace{-0.05in}

In order for the model to learn well, it is important that input and output sounds are temporally aligned.
While the Replay-NVAS data is already synchronised based on the clapper sound, due to the finite speed of sound, the sounds emitted from different locations may still arrive \CAnew{at microphones} with a delay slightly different from the one of the clapper, causing misalignments that affect training.

To align source and target audio for training, we find the delay $\tau$ that maximizes the generalized cross-correlation:
\vspace{-0.1in}
\begin{equation}
\mathcal{R}_{A_S, A_T}(\tau) = {\mathop{\mathbb{E}}}_t  [h_S(t) \cdot h_T(t - \tau)],
\vspace{-0.05in}
\end{equation}
where $h_S$ and $h_T$ are the feature embedding for $A_S$ and $A_T$ respectively at time $t$. 
We use the feature extractor $h$ from the generalized cross-correlation phase transform (GCC-PHAT) algorithm~\cite{gcc_phat}, which whitens the audio by dividing by the magnitude of the cross-power spectral density. After computing $\tau$, we shift the prediction $A_O$ by $\tau$ samples to align with the $A_T$ and obtain $A_L$.  Note that alignment is already exact for SoundSpaces-NVAS.

%\vspace{-0.05in}
\subsection{Loss}
%\vspace{-0.05in}
To compute the loss, we first encode the audio with the short-time Fourier transform (STFT), a complex-valued matrix representation of the audio where the $y$ axis represents frequency and the $x$ axis is time. 
We then compute the magnitude of the STFT,
and optimize the L1 loss between the the predicted and ground truth magnitudes as follows:
\vspace{-0.05in}
\begin{equation}
    % L = ||\textrm{abs}(STFT(A_L)) - \textrm{abs}(STFT(A_T'))||_{1} 
    L = \big| ||\textit{STFT}(A_L)||_2 - ||\textit{STFT}(A_T')||_2 \big|,
\vspace{-0.05in}
\end{equation}
where $A_T'$ is the primary sound separated from $A_T$ with $\mathcal{P}(\cdot)$.
\CA{By taking the magnitude, we do not model the exact phase values, which we find hinders learning if being included in the loss.}
\CA{See implementation details in Supp.}
%\vspace{-0.05in}
\section{Experiments} \label{sec:experiment}
%\vspace{-0.05in}
We compare with several traditional and learning-based baselines and show that ViGAS outperforms them in both a quantitative evaluation and a human subject study.

\begin{table*}[t]
% \footnotesize
\centering
\setlength{\tabcolsep}{8pt}
\begin{tabular}{c|c|c|c|c|c|c|c|c|c}
\toprule
    & \multicolumn{6}{c|}{\textbf{SoundSpaces-NVAS}} & \multicolumn{3}{c}{\textbf{Replay-NVAS}} \\
    & \multicolumn{3}{c|}{\textit{Single Environment}} & \multicolumn{3}{c|}{\textit{Novel Environment}} & \multicolumn{3}{c}{\textit{Single Environment}} \\
            & Mag & LRE & RTE & Mag & LRE & RTE & Mag & LRE & RTE \\
\midrule
 Input audio                 & 0.225 & 1.473 & 0.032 & 0.216 & 1.408 & 0.039 & 0.159 & 1.477 & 0.046  \\ 
 TF Estimator~\cite{wiener}  & 0.359 & 2.596 & 0.059 & 0.440 & 3.261 & 0.092 & 0.327 & 2.861 & 0.147 \\ 
 DSP~\cite{cheng2001introduction} & 0.302 & 3.644 & 0.044 & 0.300 & 3.689 & 0.047 & 0.463 & 1.300 & 0.067\\ 
 VAM~\cite{chen2022vam}      & 0.220 & 1.198 & 0.041 & 0.235 & 1.131 & 0.051 & 0.161 & 0.924 & 0.070 \\ 
\midrule
 ViGAS w/o visual            & 0.173 & 0.973 & 0.031 & 0.181 & 1.007 & 0.036 & 0.146	& 0.877 & \textbf{0.046} \\ 
 ViGAS                       & \textbf{0.159} & \textbf{0.782} & \textbf{0.029} & \textbf{0.175} & \textbf{0.971} & \textbf{0.034} & \textbf{0.142}	& \textbf{0.716} & 0.048 \\ 
\bottomrule
\end{tabular}
\vspace{-0.05in}
\caption{\textbf{Results on SoundSpaces-NVAS and Replay-NVAS.}
\CAnew{We report the magnitude spectrogram distance (Mag), left-right energy ratio error (LRE), and RT60 error (RTE).}
\CA{Replay-NVAS does not have novel environment setup due to data being collected in a single environment.}
For all metrics, lower is better. 
\CA{In addition to baselines, we also evaluate ViGAS w/o visual by removing the active speaker localization and visual features. 
Note that reverberation time is mostly invariant of the receiver location in the same room and thus input audio has low RTE. 
A good model should preserve this property while synthesizing the desired acoustics for the target viewpoint.
}
}
\vspace{-0.1in}
\label{tab:main}
\end{table*}

\paragraph{Evaluation.}
We measure performance from three aspects: 1. closeness to GT as measured by the \textbf{magnitude spectrogram distance (Mag)}. 2. correctness of the spatial sound as measured by the \textbf{left-right energy ratio error (LRE)}, i.e., the difference of ratio of energy between left and right channels and 3. correctness of the acoustic properties measured by \textbf{RT60 error (RTE)}~\cite{singh_image2reverb_2021,chen2022vam}, i.e., the error in reverberation time decaying by 60dB (RT60). 
\CA{We use a pretrained model~\cite{chen2022vam} to estimate RT60 directly from speech.
}

\paragraph{Baselines.}
We consider the following baselines:
1.  \textbf{Input audio}. Copying the input to the output.
2. \textbf{TF Estimator~\cite{wiener} + Nearest Neighbor}, i.e. storing the transfer function estimated during training and retrieving the nearest neighbor during test time. We estimate transfer functions with a Wiener filter~\cite{wiener} and index them with the ground-truth locations of the speaker, source viewpoint, and target viewpoint for the single environment setup and their relative pose for the novel environment setup. At test time, this method searches the database to find the nearest transfer function and applies it on the input audio. 
3. \textbf{Digital Signal Processing (DSP)~\cite{cheng2001introduction}} approach that takes the distance, azimuth, and elevation of the sound source, applies an inverse a head-related transfer function (HRTF) to estimate the speech spoken by the speaker and then applies another HRTF to estimate the audio at the target microphone location. This baseline 
adjusts the loudness of the left and right channels based on where the speaker is in the target view. We supply GT coordinates for SoundSpaces-NVAS and speakers' head positions estimated with triangulation on Replay-NVAS. 
4. \textbf{Visual Acoustic Matching (VAM)~\cite{chen2022vam}}, recently proposed for a related task of matching acoustics of input audio with a target image. This task only deals with single viewpoint and single-channel audio. We adapt their model with minimal modification by feeding in the image from the \emph{source} viewpoint and concatenating the position offset of the target microphone at the multimodal fusion step.   See Supp.~for details. 

%\vspace{-0.05in}
\subsection{Results on SoundSpaces-NVAS}
%\vspace{-0.05in}
Table~\ref{tab:main} shows the results.
For synthetic data, we consider two evaluation setups: 
1. single environment: train and test on the same environment and 2. novel environment: train and test on multiple non-overlapping Gibson environments (90/10/20 for train/val/test).

In the single environment setup, our model largely outperforms all baselines as well as our audio-only ablation on all metrics. 
TF Estimator performs poorly despite being indexed by the ground truth location values because estimating a transfer function directly from two audio clips is non-trivial and noisy for low-energy parts of the signal.
DSP also performs badly despite having the ground truth 3D coordinates of the sound source.  This is because head related transfer functions are typically recorded in anechoic chambers, which does not account for \CAnew{acoustics of  different environments, e.g., reverberation.}
Both traditional approaches perform worse than simply copying the input audio, indicating that learning-based models are needed for this challenging task. 
The recent model VAM~\cite{chen2022vam} performs much better compared to the traditional approaches but still underperforms our model. 
There is a significant difference between ViGAS w/o visual and the full model; this shows that the visual knowledge about the speaker location and the environment is important for this task. 

\begin{figure*}[t]
    \centering
    \includegraphics[width=\linewidth]{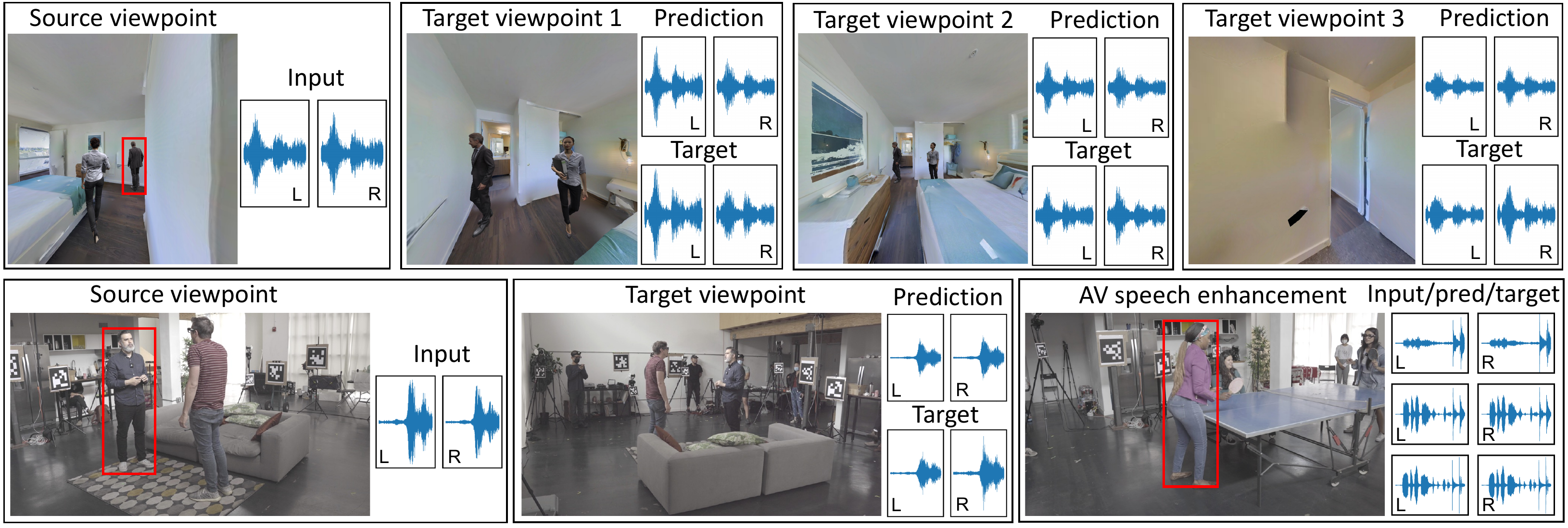}
    \vspace{-0.25in}
    \caption{\textbf{Qualitative examples}. \CA{For all binaural audio, we show the left-channel and the right-channel waveforms side-by-side.}
    Row 1: SoundSpaces-NVAS example where given the source viewpoint and input audio, the model synthesizes audio for three different target viewpoints (target views are for reference only). 
    In this case, the active speaker is the male speaker as indicated by the bounding box. 
    For target viewpoint 1, the view rotates about 90 degrees and the male speaker is on the left side and the predicted left channel is louder than the right channel. 
    Viewpoint 2 moves away from the speaker and thus yields lower amplitude compared to the first prediction. 
    For target viewpoint 3, it is completely located outside of the living room, in which case, the sound could only come from the door open on the right (louder right channel) and the reverberation also greatly increases due to the vanishing direct sound.
    Row 2: Replay-NVAS example where the speaker is located on the left in the source viewpoint which becomes the right and further from the camera in target viewpoint 2, the model also predicts lower amplitude and louder right channel. 
    On the right side, we show an example of the audio-visual speech enhancement for the active speaker. The model enhances the speech to largely match with the near-range audio (target). 
    }
    \label{fig:qual}
    \vspace{-0.1in}
\end{figure*}

Fig.~\ref{fig:qual} shows an example where given the same input source viewpoint, our model synthesizes audio for three different target viewpoints. The model reasons about how the geometry and speaker locations changes based on the source view and the target pose, and predicts the acoustic difference accordingly. See Supp.~video to listen to sounds.

For the novel environment setup, our model again outperforms all baselines.
Compared to ViGAS in the single environment setup, both the magnitude spectrogram distance and the left-right energy ratio error increase. 
This is expected because for novel (unseen) environments, single images capture limited  geometry and acoustic information.
The model fails sometime when there is a drastic viewpoint change, e.g., target viewpoint 3 in Fig.~\ref{fig:qual}. 
\CA{This setup requires the model to reason or ``imagine'' the environment based on single audio-visual observation, which poses great challenge for NVAS as well as NVS, where typically synthesis is performed in a fully observed environment.}

\begin{table}[]
    \centering
    \begin{tabular}{c|c|c|c|c}
    \toprule
              & \multicolumn{2}{c|}{\textbf{SS-NVAS}} & \multicolumn{2}{c}{\textbf{Replay-NVAS}} \\
        ViGAS & Mag & LRE & Mag & LRE \\
    \midrule
        full model & \textbf{0.159} & 0.782 & \textbf{0.142} & 0.716 \\
        w/o visual features & 0.171 & 0.897 & 0.146 & 0.920 \\
        w/o ASL & 0.161 & 0.814 & 0.143 & 0.757 \\
        w/o alignment & 0.176 & \textbf{0.771} & 0.144 & \textbf{0.706} \\
        w/o separation & 0.165 & 0.840 & 0.182 & 0.859 \\
    \bottomrule
    \end{tabular}
    \vspace{-0.05in}
    \caption{\textbf{Ablations of the model on both datasets}.}
    \label{tab:ablations}
    \vspace{-0.25in}
\end{table}

\paragraph{Ablations.}
Table~\ref{tab:ablations} shows ablations on the model design.
To understand if the model uses visual information, we ablate the visual features $V_F$ and the active speaker feature $V_L$. 
Removing the active speaker feature leads to less damage on the model performance, because without the explicitly localized active speaker, the model can still implicitly reason about the active speaker location based on the image and audio.  
If both are removed (``ViGAS w/o visual" in Table~\ref{tab:main}), the performance suffers most.

To study the effectiveness of the temporal alignment and ambient sound separation modules, we ablate them separately.
Removing the temporal alignment leads to higher Mag error and slightly lower LRE. 
As for ambient sound separation, 
the results show that optimizing for the high-energy noise-like ambient sound degrades the performance.

%\vspace{-0.05in}
\subsection{Results on Replay-NVAS}
%\vspace{-0.05in}
Table~\ref{tab:main} (right) shows the Replay-NVAS results.
Compared to SoundSpaces-NVAS, the magnitudes of all errors are smaller because 
there are less drastic acoustic changes between viewpoints (8 DLSR cameras form a circle around the participants). 
Traditional approaches like TF Estimator and DSP still perform poorly despite using the 3D coordinates of the camera and the speaker (triangulated from multiple cameras). 
VAM performs better due to end-to-end learning; however, 
our model outperforms it.
Compared to ViGAS w/o visual, the full model has much lower left-right energy ratio error \CA{and slightly higher reverberation time error}, showing that the model \CAnew{takes into account the speaker position and viewpoint change} for synthesizing the audio. 

Fig.~\ref{fig:qual} (row 2, left) shows a qualitative example. In the source viewpoint, the active speaker is on the left, while in the target viewpoint, he is further from the camera and on the right. 
\CA{The model synthesizes an audio waveform that captures the \CAnew{corresponding} acoustic change, showing that our model successfully learns from real videos.}

\paragraph{Audio-visual speech enhancement.}
In some real-world applications, e.g., hearing aid devices, the goal is to obtain the enhanced \CAnew{clean} speech of the active speaker.  This can be seen as a special case of NVAS, where the target viewpoint is the active speaker. 
Our model is capable of performing audio-visual speech enhancement without any modification.
We simply set the target audio to the near-range audio recording for the active speaker.
We show the results in Table~\ref{tab:speech_enhancement}. 
Our model obtains cleaner audio compared to the input audio (example in Fig.~\ref{fig:qual}, row 2, right).

\begin{table}[]
    \centering
\setlength{\tabcolsep}{10pt}
    \begin{tabular}{c|c|c}
    \toprule
                 & Mag & RTE \\
    \midrule
        Input    & 0.279 & 0.376\\
        ViGAS (ours)    & \textbf{0.234} & \textbf{0.122}\\
    \bottomrule
    \end{tabular}
    \vspace{-0.05in}
    \caption{\textbf{Speech enhancement on Replay-NVAS}.}
    \label{tab:speech_enhancement}
    \vspace{-0.25in}
\end{table}

\paragraph{Human subject study.} 
\CA{To supplement the quantitative metrics and evaluate how well our synthesized audio captures the acoustic change between viewpoints, we conduct a human subject study. 
We show participants the image of the target viewpoint $V_T$ as well as the audio $A_T$ as reference. 
We provide three audio samples: the input, the prediction of ViGAS, and the prediction of DSP (the most naturally sounding baseline) and ask them to select a clip that sounds closest to the target audio. 
We select 20 examples from SoundSpaces-NVAS and 20 examples from Replay-NVAS and invite 10 participants to perform the study.}

\begin{table}[]
    \centering
    \begin{tabular}{c|c|c|c}
    \toprule
    Dataset     & Input & DSP & ViGAS \\
    \midrule
        SoundSpaces-NVAS      & 24\% & 2\% & \textbf{74\%} \\
        Replay-NVAS           & 43\% & 6\% & \textbf{51\%} \\
    \bottomrule
    \end{tabular}
    \vspace{-0.05in}
    \caption{\textbf{Human Study}. Participants favor our approach over the two most realistic sounding baselines, (1) copying the input signal, and (2) a digital signal processing baseline.}
    \label{tab:user_study}
    \vspace{-0.2in}
\end{table}

See Table~\ref{tab:user_study} for the results. On the synthetic dataset SoundSpaces-NVAS, our approach is preferred over the baselines by a large margin.
This margin is lower on the real-world Replay-NVAS dataset but is still significant.

% \vspace{-0.05in}
\section{Conclusion}
% \vspace{-0.05in}
We introduce the challenging novel-view acoustic synthesis task 
and a related benchmark in form of both real and synthetic datasets. 
We propose a neural rendering model that learns to transform the sound from the source viewpoint to the target viewpoint by reasoning about the observed audio and visual stream. Our model surpasses all baselines on both datasets. We believe this research unlocks many potential applications and research in multimodal novel-view synthesis. 
In the future, \CA{we plan to incorporate active-speaker localization model into the approach and let the model jointly learn to localize and synthesize.}
% \input{sections/supp.tex}

% \newpage

{\small\bibliographystyle{ieee_fullname}%
\bibliography{egbib,vedaldi_specific,vedaldi_general}}

\begin{thebibliography}{10}\itemsep=-1pt

\bibitem{Afouras18}
Triantafyllos Afouras, Joon~Son Chung, and Andrew Zisserman.
\newblock The conversation: Deep audio-visual speech enhancement.
\newblock In {\em INTERSPEECH}, 2018.

\bibitem{lorenzo-nips2020}
Humam Alwassel, Dhruv Mahajan, Bruno Korbar, Lorenzo Torresani, Bernard Ghanem,
  and Du Tran.
\newblock Self-supervised learning by cross-modal audio-video clustering.
\newblock In {\em NeurIPS}, 2020.

\bibitem{acoustic1979}
M. Berouti, Richard Schwartz, and John Makhoul.
\newblock Enhancement of speech corrupted by acoustic noise.
\newblock In {\em IEEE International Conference on Acoustics, Speech, and
  Signal Processing}, 1979.

\bibitem{room_acoustics_review}
J.S. Bradley.
\newblock Review of objective room acoustics measures and future needs.
\newblock {\em Applied Acoustics}, 2011.

\bibitem{Matterport3D}
Angel Chang, Angela Dai, Thomas Funkhouser, Maciej Halber, Matthias Niessner,
  Manolis Savva, Shuran Song, Andy Zeng, and Yinda Zhang.
\newblock Matterport3d: Learning from rgb-d data in indoor environments.
\newblock {\em 3DV}, 2017.
\newblock MatterPort3D dataset license available at:
  \url{http://kaldir.vc.in.tum.de/matterport/MP_TOS.pdf}.

\bibitem{chen2022vam}
Changan Chen, Ruohan Gao, Paul Calamia, and Kristen Grauman.
\newblock Visual acoustic matching.
\newblock In {\em CVPR}, 2022.

\bibitem{chen22soundspaces2}
Changan Chen, Carl Schissler, Sanchit Garg, Philip Kobernik, Alexander Clegg,
  Paul Calamia, Dhruv Batra, Philip~W Robinson, and Kristen Grauman.
\newblock Soundspaces 2.0: A simulation platform for visual-acoustic learning.
\newblock In {\em NeurIPS 2022 Datasets and Benchmarks Track}, 2022.

\bibitem{chen_dereverb23}
Changan Chen, Wei Sun, David Harwath, and Kristen Grauman.
\newblock Learning audio-visual dereverberation.
\newblock In {\em ICASSP}, 2023.

\bibitem{cheng2001introduction}
Corey~I. Cheng and Gregory~H. Wakefield.
\newblock Introduction to head-related transfer functions (hrtfs):
  representations of hrtfs in time, frequency, and space.
\newblock {\em journal of the audio engineering society}, 49(4):231--249, april
  2001.

\bibitem{mmtrack2020}
MMTracking Contributors.
\newblock {MMTracking: OpenMMLab} video perception toolbox and benchmark.
\newblock \url{https://github.com/open-mmlab/mmtracking}, 2020.

\bibitem{defossez2020real}
Alexandre Defossez, Gabriel Synnaeve, and Yossi Adi.
\newblock Real time speech enhancement in the waveform domain.
\newblock In {\em Interspeech}, 2020.

\bibitem{eaton20161ACE}
James Eaton, Nikolay Gaubitch, Allistair Moore, and Patrick Naylor.
\newblock Estimation of room acoustic parameters: The {ACE} challenge.
\newblock {\em IEEE/ACM Transactions on Audio, Speech, and Language
  Processing}, 24(10), 2016.

\bibitem{subspace1995}
Yariv Ephraim and Harry L.~Van Trees.
\newblock A signal subspace approach for speech enhancement.
\newblock {\em IEEE Transactions on Speech and Audio Processing}, 1995.

\bibitem{ephrat2018looking}
Ariel Ephrat, Inbar Mosseri, Oran Lang, Tali Dekel, Kevin Wilson, Avinatan
  Hassidim, William~T Freeman, and Michael Rubinstein.
\newblock Looking to listen at the cocktail party: A speaker-independent
  audio-visual model for speech separation.
\newblock In {\em SIGGRAPH}, 2018.

\bibitem{fu2019metricGAN}
Szu-Wei Fu, Chien-Feng Liao, Yu Tsao, and Shou-De Lin.
\newblock Metricgan: Generative adversarial networks based black-box metric
  scores optimization for speech enhancement.
\newblock In {\em International Conference on Machine Learning (ICML)}, 2019.

\bibitem{gamper2018blind}
Hannes Gamper and Ivan~J Tashev.
\newblock Blind reverberation time estimation using a convolutional neural
  network.
\newblock In {\em 2018 16th International Workshop on Acoustic Signal
  Enhancement (IWAENC)}, pages 136--140, 2018.

\bibitem{25d-visual-sound}
Ruohan Gao and Kristen Grauman.
\newblock 2.5d visual sound.
\newblock In {\em CVPR}, 2019.

\bibitem{resnet18}
Kaiming He, Xiangyu Zhang, Shaoqing Ren, and Jian Sun.
\newblock Deep residual learning for image recognition.
\newblock In {\em CVPR}, pages 770--778, 2016.

\bibitem{hu-localize-nips2020}
Di Hu, Rui Qian, Minyue Jiang, Xiao Tan, Shilei Wen, Errui Ding, Weiyao Lin,
  and Dejing Dou.
\newblock Discriminative sounding objects localization via self-supervised
  audiovisual matching.
\newblock In {\em NeurIPS}, 2020.

\bibitem{jain21putting}
Ajay Jain, Matthew Tancik, and Pieter Abbeel.
\newblock Putting nerf on a diet: Semantically consistent few-shot view
  synthesis.
\newblock In {\em Proc. {ICCV}}, 2021.

\bibitem{hao2022asl}
Hao Jiang, Calvin Murdock, and Vamsi~Krishna Ithapu.
\newblock Egocentric deep multi-channel audio-visual active speaker
  localization.
\newblock In {\em CVPR}, 2022.

\bibitem{klein2019real}
Florian Klein, Annika Neidhardt, and Marius Seipel.
\newblock Real-time estimation of reverberation time for selection of suitable
  binaural room impulse responses.
\newblock In {\em Audio for Virtual, Augmented and Mixed Realities: Proceedings
  of 5th International Conference on Spatial Audio (ICSA)}, pages 145--150,
  2019.

\bibitem{gcc_phat}
C. Knapp and G. Carter.
\newblock The generalized correlation method for estimation of time delay.
\newblock {\em IEEE Transactions on Acoustics, Speech, and Signal Processing},
  1976.

\bibitem{korbar-nips2018}
Bruno Korbar, Du Tran, and Lorenzo Torresani.
\newblock Cooperative learning of audio and video models from self-supervised
  synchronization.
\newblock In {\em NeurIPS}, 2018.

\bibitem{kulhanek22viewformer:}
Jon{\'a}{\v s} Kulh{\'a}nek, Erik Derner, Torsten Sattler, and Robert Babu{\v
  s}ka.
\newblock {ViewFormer}: {NeRF-free} neural rendering from few images using
  transformers.
\newblock In {\em Proc. {ECCV}}, 2022.

\bibitem{room_acoustics}
Heinrich Kuttruff.
\newblock {\em Room Acoustics}.
\newblock Boca Raton, 6th edition, 2016.

\bibitem{li20neural}
Zhengqi Li, Simon Niklaus, Noah Snavely, and Oliver Wang.
\newblock Neural scene flow fields for space-time view synthesis of dynamic
  scenes.
\newblock {\em arXiv.cs}, abs/2011.13084, 2020.

\bibitem{liu21neural}
Lingjie Liu, Marc Habermann, Viktor Rudnev, Kripasindhu Sarkar, Jiatao Gu, and
  Christian Theobalt.
\newblock Neural humans: Pose-controlled free-view synthesis of human actors
  with template-guided neural radiance fields.
\newblock In {\em arXiv}, 2021.

\bibitem{luo2022learning}
Andrew Luo, Yilun Du, Michael~J Tarr, Joshua~B Tenenbaum, Antonio Torralba, and
  Chuang Gan.
\newblock Learning neural acoustic fields.
\newblock In {\em NeurIPS}, 2022.

\bibitem{mack2020single}
Wolfgang Mack, Shuwen Deng, and Emanu{\"e}l~AP Habets.
\newblock Single-channel blind direct-to-reverberation ratio estimation using
  masking.
\newblock In {\em INTERSPEECH}, pages 5066--5070, 2020.

\bibitem{majumder2022fewshot}
Sagnik Majumder, Changan Chen, Ziad Al-Halah, and Kristen Grauman.
\newblock Few-shot audio-visual learning of environment acoustics.
\newblock In {\em Thirty-Sixth Conference on Neural Information Processing
  Systems}, 2022.

\bibitem{Martin-Brualla_2021_CVPR}
Ricardo Martin-Brualla, Noha Radwan, Mehdi S.~M. Sajjadi, Jonathan~T. Barron,
  Alexey Dosovitskiy, and Daniel Duckworth.
\newblock Nerf in the wild: Neural radiance fields for unconstrained photo
  collections.
\newblock In {\em Proceedings of the IEEE/CVF Conference on Computer Vision and
  Pattern Recognition (CVPR)}, pages 7210--7219, June 2021.

\bibitem{Michelsanti19}
Daniel Michelsanti, Zheng-Hua Tan, Shi-Xiong Zhang, Yong Xu, Meng Yu, Dong Yu,
  and Jesper Jensen.
\newblock An overview of deep-learning-based audio-visual speech enhancement
  and separation.
\newblock In {\em arXiv}, 2020.

\bibitem{mildenhall2020nerf}
Ben Mildenhall, Pratul~P. Srinivasan, Matthew Tancik, Jonathan~T. Barron, Ravi
  Ramamoorthi, and Ren Ng.
\newblock Nerf: Representing scenes as neural radiance fields for view
  synthesis.
\newblock In {\em ECCV}, 2020.

\bibitem{morgado-spatial-nips2020}
Pedro Morgado, Yi Li, and Nuno Vasconcelos.
\newblock Learning representations from audio-visual spatial alignment.
\newblock In {\em NeurIPS}, 2020.

\bibitem{morgado-2018}
Pedro Morgado, Nono Vasconcelos, Timothy Langlois, and Oliver Wang.
\newblock Self-supervised generation of spatial audio for 360 video.
\newblock In {\em NeurIPS}, 2018.

\bibitem{murgai2017blind}
Prateek Murgai, Mark Rau, and Jean-Marc Jot.
\newblock Blind estimation of the reverberation fingerprint of unknown acoustic
  environments.
\newblock In {\em Audio Engineering Society Convention 143}. Audio Engineering
  Society, 2017.

\bibitem{niemeyer22regnerf:}
Michael Niemeyer, Jonathan~T. Barron, Ben Mildenhall, Mehdi S.~M. Sajjadi,
  Andreas Geiger, and Noha Radwan.
\newblock {RegNeRF}: Regularizing neural radiance fields for view synthesis
  from sparse inputs.
\newblock In {\em Proc. {CVPR}}, 2022.

\bibitem{oord2016wavenet}
Aaron van~den Oord, Sander Dieleman, Heiga Zen, Karen Simonyan, Oriol Vinyals,
  Alex Graves, Nal Kalchbrenner, Andrew Senior, and Koray Kavukcuoglu.
\newblock Wavenet: A generative model for raw audio, 2016.
\newblock cite arxiv:1609.03499.

\bibitem{owens2018audio}
Andrew Owens and Alexei~A Efros.
\newblock Audio-visual scene analysis with self-supervised multisensory
  features.
\newblock In {\em ECCV}, 2018.

\bibitem{librispeech}
Vassil Panayotov, Guoguo Chen, Daniel Povey, and Sanjeev Khudanpur.
\newblock Librispeech: an asr corpus based on public domain audio books.
\newblock In {\em 2015 IEEE international conference on acoustics, speech and
  signal processing (ICASSP)}, pages 5206--5210. IEEE, 2015.

\bibitem{park20deformable}
Keunhong Park, Utkarsh Sinha, Jonathan~T. Barron, Sofien Bouaziz, Dan~B.
  Goldman, Steven~M. Seitz, and Ricardo Martin{-}Brualla.
\newblock Deformable neural radiance fields.
\newblock {\em CoRR}, abs/2011.12948, 2020.

\bibitem{pumarola20d-nerf:}
Albert Pumarola, Enric Corona, Gerard Pons{-}Moll, and Francesc
  Moreno{-}Noguer.
\newblock {D-NeRF}: Neural radiance fields for dynamic scenes.
\newblock {\em arXiv.cs}, abs/2011.13961, 2020.

\bibitem{Pumarola_2021_CVPR}
Albert Pumarola, Enric Corona, Gerard Pons-Moll, and Francesc Moreno-Noguer.
\newblock D-nerf: Neural radiance fields for dynamic scenes.
\newblock In {\em Proceedings of the IEEE/CVF Conference on Computer Vision and
  Pattern Recognition (CVPR)}, pages 10318--10327, June 2021.

\bibitem{ratnarajah2022mesh2ir}
Anton Ratnarajah, Zhenyu Tang, Rohith~Chandrashekar Aralikatti, and Dinesh
  Manocha.
\newblock Mesh2ir: Neural acoustic impulse response generator for complex 3d
  scenes.
\newblock In {\em ACM Multimedia}, 2022.

\bibitem{reizenstein21common}
Jeremy Reizenstein, Roman Shapovalov, Philipp Henzler, Luca Sbordone, Patrick
  Labatut, and David Novotny.
\newblock {Common Objects in 3D}: Large-scale learning and evaluation of
  real-life {3D} category reconstruction.
\newblock In {\em Proc. {CVPR}}, 2021.

\bibitem{richard2022deepimpulse}
Alexander Richard, Peter Dodds, and Vamsi~Krishna Ithapu.
\newblock Deep impulse responses: Estimating and parameterizing filters with
  deep networks.
\newblock In {\em IEEE International Conference on Acoustics, Speech and Signal
  Processing}, 2022.

\bibitem{richard2021binaural}
Alexander Richard, Dejan Markovic, Israel~D Gebru, Steven Krenn, Gladstone
  Butler, Fernando de~la Torre, and Yaser Sheikh.
\newblock Neural synthesis of binaural speech from mono audio.
\newblock In {\em International Conference on Learning Representations}, 2021.

\bibitem{ava_active_speaker}
Joseph Roth, Sourish Chaudhuri, Ondrej Klejch, Radhika Marvin, Andrew
  Gallagher, Liat Kaver, Sharadh Ramaswamy, Arkadiusz Stopczynski, Cordelia
  Schmid, Zhonghua Xi, and Caroline Pantofaru.
\newblock Ava-activespeaker: An audio-visual dataset for active speaker
  detection.
\newblock In {\em ICASSP}, 2020.

\bibitem{singh_image2reverb_2021}
Nikhil Singh, Jeff Mentch, Jerry Ng, Matthew Beveridge, and Iddo Drori.
\newblock Image2reverb: Cross-modal reverb impulse response synthesis.
\newblock In {\em ICCV}, 2021.

\bibitem{sitzmann19scene}
Vincent Sitzmann, Michael Zollh{\"{o}}fer, and Gordon Wetzstein.
\newblock Scene representation networks: Continuous {3D}-structure-aware neural
  scene representations.
\newblock {\em Proc. {NeurIPS}}, 2019.

\bibitem{straub2019replica}
Julian Straub, Thomas Whelan, Lingni Ma, Yufan Chen, Erik Wijmans, Simon Green,
  Jakob~J Engel, Raul Mur-Artal, Carl Ren, Shobhit Verma, et~al.
\newblock The {Replica} dataset: A digital replica of indoor spaces.
\newblock {\em arXiv preprint arXiv:1906.05797}, 2019.

\bibitem{su21a-nerf:}
Shih{-}Yang Su, Frank Yu, Michael Zollh{\"{o}}fer, and Helge Rhodin.
\newblock {A-NeRF}: Surface-free human 3d pose refinement via neural rendering.
\newblock {\em arXiv.cs}, abs/2102.06199, 2021.

\bibitem{tao2021someone}
Ruijie Tao, Zexu Pan, Rohan~Kumar Das, Xinyuan Qian, Mike~Zheng Shou, and
  Haizhou Li.
\newblock Is someone speaking? exploring long-term temporal features for
  audio-visual active speaker detection.
\newblock In {\em Proceedings of the 29th ACM International Conference on
  Multimedia}, page 3927–3935, 2021.

\bibitem{tretschk20non-rigid}
Edgar Tretschk, Ayush Tewari, Vladislav Golyanik, Michael Zollh{\"{o}}fer,
  Christoph Lassner, and Christian Theobalt.
\newblock Non-rigid neural radiance fields: Reconstruction and novel view
  synthesis of a deforming scene from monocular video.
\newblock {\em arXiv.cs}, abs/2012.12247, 2020.

\bibitem{wiener}
Norbert Wiener.
\newblock Extrapolation, interpolation, and smoothing of stationary time
  series.
\newblock {\em Report of the Services 19, Research Project DIC-6037 MIT}, 1942.

\bibitem{xiazamirhe2018gibsonenv}
Fei Xia, Amir R.~Zamir, Zhi-Yang He, Alexander Sax, Jitendra Malik, and Silvio
  Savarese.
\newblock Gibson {Env}: real-world perception for embodied agents.
\newblock In {\em Computer Vision and Pattern Recognition (CVPR), 2018 IEEE
  Conference on}. IEEE, 2018.

\bibitem{xiong2018joint}
Feifei Xiong, Stefan Goetze, Birger Kollmeier, and Bernd~T Meyer.
\newblock Joint estimation of reverberation time and early-to-late
  reverberation ratio from single-channel speech signals.
\newblock {\em IEEE/ACM Transactions on Audio, Speech, and Language
  Processing}, 27(2):255--267, 2018.

\bibitem{yu21pixelnerf:}
Alex Yu, Vickie Ye, Matthew Tancik, and Angjoo Kanazawa.
\newblock {PixelNeRF}: Neural radiance fields from one or few images.
\newblock In {\em Proc. {CVPR}}, 2021.

\bibitem{zhou19}
Hang Zhou, Ziwei Liu, Xudong Xu, Ping Luo, and Xiaogang Wang.
\newblock Vision-infused deep audio inpainting.
\newblock In {\em ICCV}, 2019.

\end{thebibliography}

\clearpage
\end{document}